\documentclass[runningheads]{llncs}

\usepackage[year=2026,ID=3425]{eccv}

\usepackage{eccvabbrv}
\usepackage[table,xcdraw]{xcolor}
\usepackage{graphicx}
\usepackage{booktabs}
\usepackage{multirow}
\usepackage{adjustbox}
\usepackage{makecell}
\usepackage{soul}

\newcommand{\JHNOTE}[1]{\textcolor{red}{[\textbf{JH Comment}: #1]}}

\newcommand{\CZADD}[1]{\textcolor{black}{ #1}}

\usepackage[accsupp]{axessibility}  

\usepackage[citecolor=blue, colorlinks]{hyperref}
\usepackage{orcidlink}
\usepackage{wrapfig,lipsum,booktabs}

\begin{document}

\title{REFINE: Super-efficient 3D Gaussian Splatting Pruning via \underline{Re}ndering-\underline{F}ree Primitive \underline{I}mporta\underline{n}c\underline{e}}

\titlerunning{REFINE}

\author{Zhang Chen\inst{1} \and
Shuai Wan\inst{1} \thanks{Corresponding author.} \and
Mengting Yu\inst{1}  \and
Fuzheng Yang\inst{2}  \and
Junhui Hou\inst{3}  }

\authorrunning{Z.~Chen et al.}

\institute{School of Electronics and Information, Northwestern Polytechnical University\\
\and
School of Telecommunication Engineering, Xidian University\\
\and
Department of Computer Science, City University of Hong Kong\\
\textcolor{magenta}{https://zhangchen2022.github.io/REFINE.github.io/}
}

\maketitle

\begin{abstract} 
Existing pruning methods for 3D Gaussian splatting (3DGS) suffer from either severe quality degradation or prohibitive computational overhead. In this paper, we propose REFINE, a highly accelerated 3DGS pruning framework centered on a novel rendering-free primitive importance metric. Our approach leverages an analytically approximated, rendering-aware Hessian field to quantify the expected perceptual error induced by the removal of individual primitives. By modeling the joint modulation of visibility, projection geometry and the content adaptive hyperparameter, we entirely bypass costly forward rendering passes and derive an anisotropic perceptual weight field that serves as a high-fidelity proxy for primitive importance. Extensive experiments across multiple benchmark datasets demonstrate that REFINE maintains highly competitive rendering quality while achieving a $3,000\times$ reduction in pruning-related computational complexity, translating to a practical $\sim 20\times$ speedup in device latency compared to state-of-the-art pruning methods.

 \and Efficiency
\end{abstract}

\section{Introduction}
\label{sec:intro}
3D Gaussian Splatting (3DGS) \cite{kerbl20233d} has emerged as a revolutionary representation for novel view synthesis (NVS) \cite{barron2022mipnerf360,deng2022depth}. By explicitly representing scenes using 3D Gaussian primitives and utilizing an efficient tile-based rasterizer, 3DGS achieves real-time rendering rates exceeding 100 FPS at 1080p resolution\cite{chen2025pcgs,chen2024fast}.

Despite its impressive rendering performance, the explicit representation of 3DGS is notoriously storage-heavy\cite{duisterhof2024deformgs,chen2025haifgs,chen2024hac}. To accurately capture high-frequency details and complex geometry, adaptive density control often generates millions of redundant Gaussian primitives for a single scene\cite{fang2024minisplatting,chen2025pcgs}. Since each primitive requires 59 parameters, storage requirements frequently reach the gigabyte (GB) level \cite{navaneet2024compgs,huang2025hierarchical}. This massive redundancy severely limits the practical utility of 3DGS in resource-constrained environments, such as VR/AR headsets and mobile phones, and creates significant bottlenecks for network streaming\cite{kong2025efficient,niedermayr2024compressed,papantonakis2024reducing,navaneet2024compgs}. Consequently, pruning redundant primitives while maintaining rendering quality has become a critical necessity \cite{fan2024lightgaussian}.  
Existing 3DGS pruning methods generally fall into two categories, both of which face an irreconcilable contradiction between computational efficiency and perceptual accuracy. The first category comprises parameter-based methods (e.g., \CZADD{Opacity \cite{kerbl20233d}}), which evaluate primitive importance directly in the parameter space. While computationally efficient, these methods ignore the physical non-linear mapping of the rendering pipeline, often leading to the erroneous pruning of critical high-frequency primitives and resulting in blurred rendering\cite{xie2024mesongs}. The second category includes render-aware methods (e.g., PUP 3D-GS\cite{hanson2025pup}). While these preserve perceptual accuracy, they require forward rendering to accumulate sensitivity scores, trapping them in expensive rasterization loops and resulting in extremely high processing times. \par

\begin{wrapfigure}{r}{0.45\textwidth}
  \vspace{-20pt} %
  \centering
  \includegraphics[width=\linewidth]{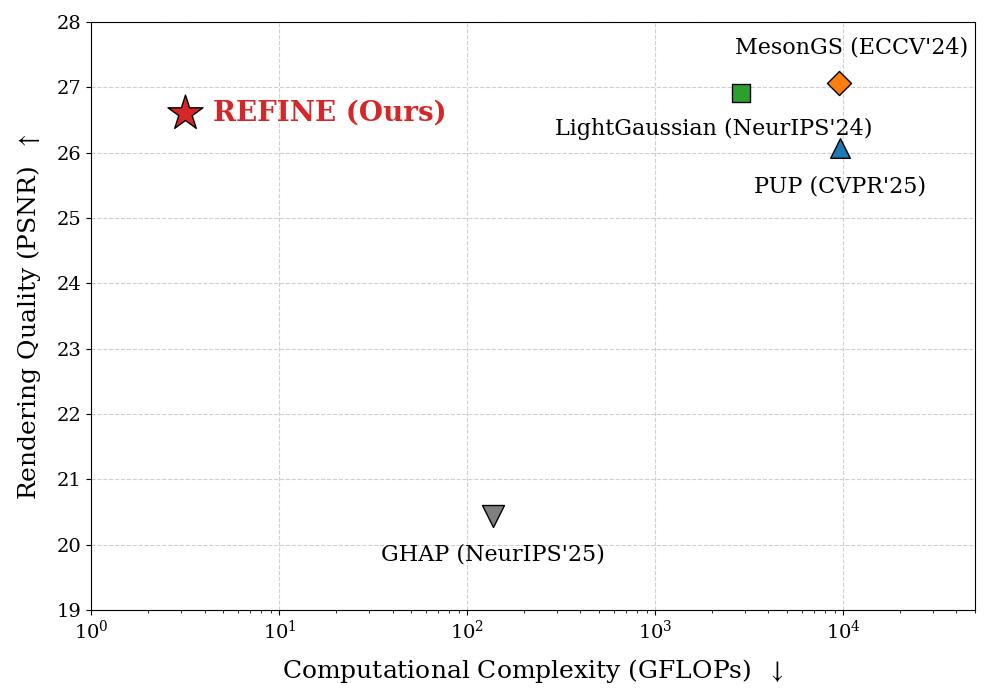}
  \caption{\CZADD{Quality vs. Efficiency Trade-off on MipNeRF360 (Ratio = 0.5).}}
  \label{fig:tradeoff}
  \vspace{-20pt} %
\end{wrapfigure}
To address the challenge, we propose REFINE, a completely rendering-free, post-processing pruning framework that achieves state-of-the-art rendering quality with \textit{exceptional} efficiency. To resolve the inconsistency between parameter space heuristics and image space visual degradation, we formalize primitive importance assessment through an analytically approximated Hessian field. Specifically, we quantify the expected visual error induced by primitive removal by decoupling the parameter gradients into two physically interpretable components: view-dependent visibility and geometric projection. Furthermore, we design the content-adaptive hyperparameter that dynamically calibrate the sensitivity of different physical attributes. By integrating these components, REFINE successfully obtains an importance score entirely without forward rendering. Extensive experiments demonstrate that our approach matches the visual fidelity of rendering-based methods while reducing the computational overhead by orders of magnitude, as shown in Fig. \ref{fig:tradeoff}.

\section{Related Work}

\noindent \textbf{Parameter-space Post-processing Pruning.}
The first category consists of methods that operate directly on the explicit geometric or appearance parameters of Gaussians. For instance, Lee et al. \cite{lee2024compact} introduce a learnable mask parameter that considers both scale and opacity together, avoiding fixed thresholds to enable end-to-end learning of pruning decisions. OMG \cite{lee2025optimized} extends significance scoring by factoring in Local Distinctiveness, which evaluates a Gaussian's importance by comparing its appearance features to its k-nearest neighbors. Alternatively, GHAP\cite{wang2025gaussian} approaches compaction from the perspective of optimal transport, formulating it as a global reduction problem of Gaussian Mixture Models to mathematically merge redundant primitives into new representative centers. While these parameter-space methods are computationally efficient, they face inherent limitations in preserving rendering fidelity.\par

\vspace{0.5em}
\noindent\textbf{Rendering-aware Post-processing Pruning.}
The second category comprises render-aware pruning methods, which aim to link primitive removal directly to rendering error. For example, \CZADD{LightGaussian\cite{fan2024lightgaussian} combines Gaussian attributes such as opacity and volume with view-dependent global significance accumulated through rasterization over training views.} EAGLES \cite{girish2024eagles} introduces a criterion to identify inefficient Gaussians by calculating their influence at a specific pixel based on alpha blending and transmittance values. MesonGS\cite{xie2024mesongs} proposes a render-aware metric that combines view-independent parameters with a view-dependent score obtained via forward rendering passes to accumulate physical pixel contributions. Going a step further into optimization quality, methods like PUP 3D-GS\cite{hanson2025pup} utilize the Fisher approximation of the Hessian matrix to calculate the sensitivity of reconstruction error to each Gaussian spatial parameter. Speedy-Splat \cite{hanson2025speedy} reparameterizes this Hessian approximation to further reduce storage requirements during the pruning process. Additionally, Trimming\cite{ali2024trimming} and ELMGS \cite{ali2025elmgs} propose Gradient-aware Pruning, which uses both opacity and gradient signals to prune inefficient Gaussians. Although these methods address the perceptual accuracy issues of parameter-space approaches, their reliance on the rendering pipeline introduces significant overhead. These render-aware methods typically require forward rendering, and in some cases backpropagation, resulting in extremely high computational complexity and processing time.

\vspace{0.5em}
\noindent \textbf{3DGS Compression.}
To address the massive memory demands of 3DGS \cite{youn2025success}, recent compression methods extend beyond pruning into parameter and restructuring compression\cite{chen2026feedforward}. Parameter compression minimizes storage via attribute quantization (e.g., CompGS \cite{liu2024compgs}, Niedermayr et al. \cite{niedermayr2024compressed}) and entropy coding (e.g., HAC \cite{chen2024hac}). Alternatively, restructuring methods fundamentally modify the 3DGS architecture for compactness, employing sparse anchors (e.g., Scaffold-GS \cite{lu2024scaffoldgs}), neural MLPs (e.g., EAGLES \cite{girish2024eagles}), or geometric structures like Octrees (e.g., Octree-GS \cite{ren2025octreegs}).

In summary, current post-processing pruning methods present a forced choice between the inaccuracy of parameter-space heuristics and the computational burden of image-space rendering evaluation. Our proposed REFINE bridges this gap by formulating primitive importance assessment directly on the parameter manifold, achieving the accuracy of render-aware methods with the efficiency of heuristic rules.\par

\section{Proposed Method}
Our goal is to quantify the visual importance of each Gaussian primitive \textit{without} additional, costly forward rendering, which is then used to guide pruning. To achieve this, we first analyze the parameter space Hessian matrix to establish an importance weight field (Section \ref{sec3.1}). Next, we analytically model these Hessian weights and calculate primitive-wise importance scores (Section \ref{sec3.2}). Finally, we execute highly efficient pruning using the importance scores (Section \ref{sec 3.3}).

\subsection{Preliminary} \label{sec3.1}
\vspace{0.5em}
\noindent\textbf{3D Gaussian Splatting.} 3DGS \cite{kerbl20233d} is a point-based NVS technique that uses 3D Gaussians to model the scene. Formally, we represent a scene as a set of $N$ Gaussian primitives and define the set as: 
\begin{equation}
{\cal G} = \left\{ {{G_i} = ({\mu _i},{s_i},{q_i},{c_i},{\alpha _i})} \right\}_{i = 1}^N ,
\end{equation}
where $\mu_i \in \mathbb{R}^3$ denotes the center position (mean), $s_i \in \mathbb{R}^{3}$ represents the scaling factors, and $q_i \in \mathbb{R}^4$ is the rotation quaternion. To capture view-dependent appearance, $c_i \in \mathbb{R}^{3 \times 16}$ denotes the Spherical Harmonic (SH) coefficients, while $\alpha_i \in \mathbb{R}$ represents the opacity of the primitive \cite{lei2025mosca,jiang2024hifi4g}.

Given a camera pose, a differentiable rasterizer renders a 2D image of resolution $h\times w$, denoted as $ \mathcal{I}\in\mathbb{R}^{h\times w\times 3}$, by projecting the 3D Gaussian primitives. 
Denote by $\mathbf{I} \in \mathbb{R}^{h \times w \times 3}$ the flattened rendered image.
We formalize the 3DGS rendering process as a differentiable mapping $R (\cdot): \mathcal{M} \to \mathcal{I}$, such that $\mathbf{I} = R (\mathcal{G})$. This operator represents a complex, non-linear transformation from the high-dimensional Gaussian parameter space 
$\mathcal{M}$ to the rendered 2D image space $\mathcal{I}$.  

\vspace{0.5em}
\noindent\textbf{Importance Modeling via Scaled Fisher Information.} \label{sec3.2}
When the primitives in $\mathcal{G}$ are subject to a perturbation $\Delta \mathcal{G}$ (via primitive pruning), we analytically approximate the resulting image space variation using a first-order Taylor expansion:
\begin{equation}
\Delta \mathbf{I} \approx \mathbf{J}_{{R}} \cdot \Delta \mathcal{G},
  \label{eq:2}
\end{equation}
where $\mathbf{J}_{{R}} = \frac{\partial {R}}{\partial \mathcal{G}}$ denotes the Jacobian matrix of the rendering function ${R} (\cdot)$ with respect to the primitives $\mathcal{G}$. Given that quality assessment in image space is commonly characterized by the $L_2$ norm, we project this metric into the parameter space via $Q (\cdot)$
\begin{equation}
Q({\Delta}\mathcal{ G}) = \left\| {\Delta {\bf{I}}} \right\|_2^2 \approx \Delta {\mathcal{G}^ \top }(\underbrace {{\bf{J}}_R^ \top {{\bf{J}}_R}}_{\bf{H}})\Delta \mathcal{G},
  \label{eq:3}
\end{equation}
where $Q$ quantifies the image quality degradation caused by pruning, directly reflecting the importance of the removed primitives. The term $\mathbf{H} = \mathbf{J}_{{R}}^\top \mathbf{J}_{{R}}$ constitutes the Gauss-Newton approximation of the Hessian matrix\cite{foresee1997gauss}, defining a metric tensor that characterizes the importance of the parameter space. This formulation possesses a rigorous statistical interpretation: assuming the observed image $\mathbf{I}_{obs}$ is corrupted by independent and identically distributed Gaussian white noise $\epsilon \sim \mathcal{N}(0, \sigma^2 \mathbf{I})$, i.e., $\mathbf{I}_{obs} = {R}(\mathcal{G}) + \epsilon$, where $\sigma$ is the standard deviation of the noise, the negative log-likelihood of the rendering is given by \cite{fujita2022fisher}:
\begin{equation}
-\log \phi(\mathbf{I}_{obs} | \mathcal{G}) \propto \frac{1}{2\sigma^2} \| \mathbf{I}_{obs} - {R}(\mathcal{G}) \|_2^2,
  \label{eq:4}
\end{equation}
where $\phi$ denotes the likelihood function. The Fisher Information Matrix (FIM) $\mathbf{F}$, defined as the covariance of the gradient of the log-likelihood \cite{fujita2022fisher, liu2020quantum}, simplifies under this Gaussian assumption to:
\begin{equation}
\mathbf{F} = \mathbb{E} \left[ \left( \frac{\partial \log \phi}{\partial \mathcal{G}} \right) \left( \frac{\partial \log \phi}{\partial \mathcal{G}} \right)^\top \right] \approx \frac{1}{\sigma^2} \mathbf{J}_{{R}}^\top \mathbf{J}_{{R}} .
\label{eq:5}
\end{equation}
Consequently, $\mathbf{H}$ in Eq. (\ref{eq:3}) is essentially a scaled FIM. It directly quantifies the information contribution of each primitive to the final rendered image. 

\subsection{Rendering-free Primitive Importance Metric}\label{sec3.2}
The 3D Gaussian representation of a scene $\mathcal{G}$ typically contains over a million primitives, making the calculation and storage of the dense matrix $\mathbf{H}$ infeasible. To this end, we introduce two structural assumptions:  (\textbf{1}) \textit{Primitive Independence}: ignoring interaction between different Gaussian primitives; and (\textbf{2}) \textit{Attribute Orthogonality}: ignoring second-order couplings between different attributes within the same primitive. See the experiments in Section \ref{sec5.4} for the rationality verification.

Based on these assumptions, $\mathbf{H}$ can be further approximated as a diagonal matrix, i.e., $\mathbf{W} = \text{diag}(\mathbf{H})$ with the $i$-th diagonal element being $w_{i}$. 
By treating the perturbation $\Delta \mathcal{G}$ as the removal of a primitive ${G_i}$ and expanding this variation across its different attributes, we can express the importance of the $i$-th primitive as: 
\begin{equation}
D({G_i}) = \sum\limits_{k \in \{ gem, col, opa \}} {w_i^k \cdot \tilde G_i^k} ,
\label{eq:7}
\end{equation}
where $\tilde G_i^k$ denotes the $k$-th attribute subset (geometry area $gem$, color $col$, and opacity $opa$) of the $i$-th Gaussian primitive $G_i$, and $w_{i}^{k}$ represents the corresponding Hessian weight of $G_{i}^{k}$. 

\begin{figure}[tb]
  \centering
  \includegraphics[height=5.5cm]{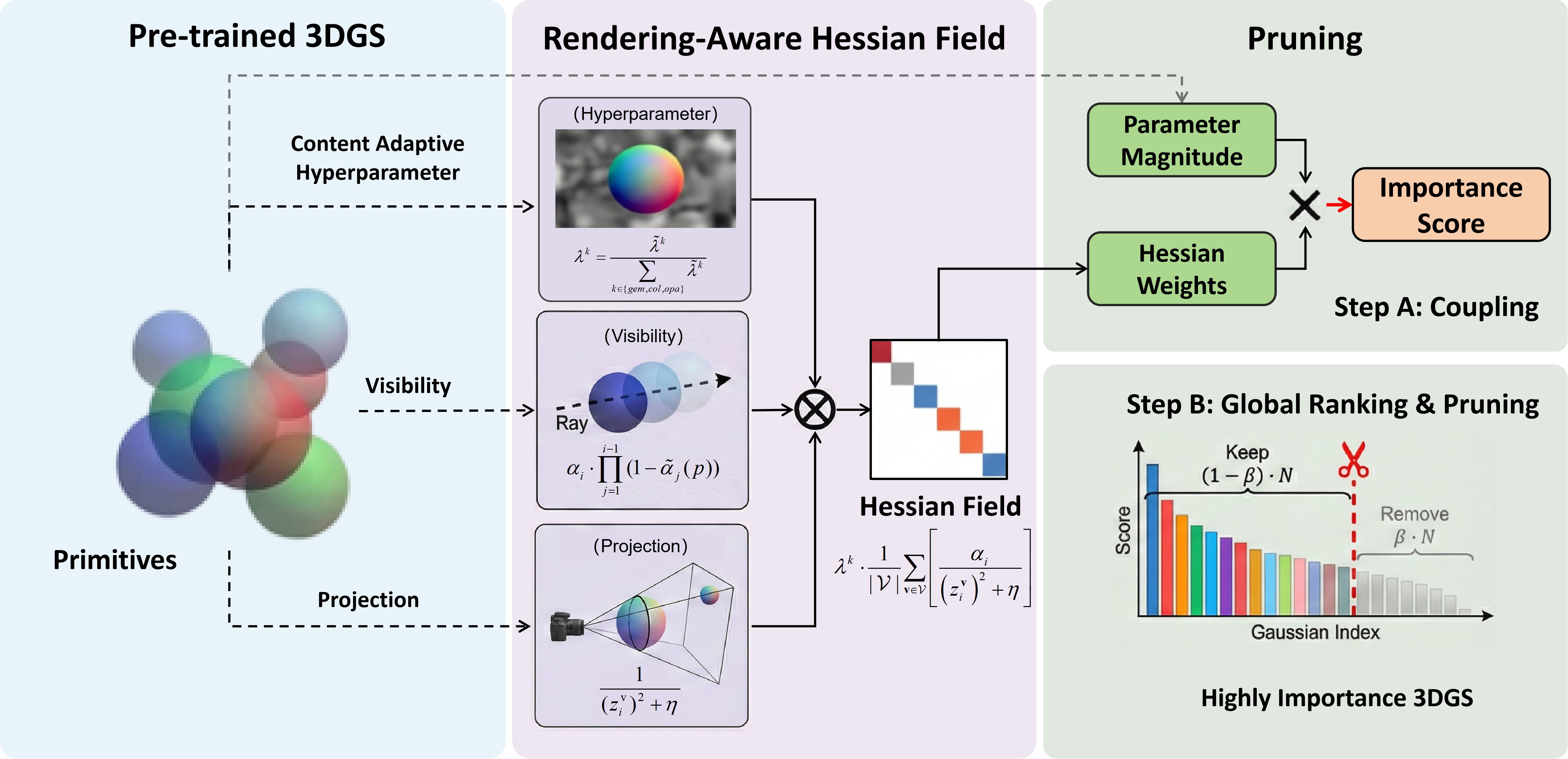}
  \caption{
    Overview of our \textbf{REFINE}, a super-efficient and effective pruning method for 3DGS.    
    \textbf{(Middle)} The rendering-aware Hessian field is computed by decomposing into visibility and projection \textit{without} performing feedforward rendering. 
    \textbf{(Right)} The importance score is computed by coupling parameter magnitudes with Hessian weights. And primitives are globally ranked by this score and pruned by the ratio value, achieving super-efficient pruning without any rendering passes.
}
  \label{fig2}
\end{figure}


Recall that $w_{i}^{k}$ is essentially the squared magnitude of the corresponding column vector in the rendering Jacobian matrix. Consequently, $w_{i}^{k}$ can be calculated by the expected squared $L_2$-norm of the image error gradient induced by this parameter across a set of sampled camera viewpoints $\mathcal{V}$:
\begin{equation}
w_{i}^{k} = \mathbb{E}_{\mathbf{v} \in \mathcal{V}} \left[ \left\| \frac{\partial {R}^{\mathbf{v}}}{\partial G_{i}^{k}} \right\|_2^2 \right] ,
  \label{eq:8}
\end{equation}
where $\mathbb{E}[\cdot]$ denotes the mathematical expectation. Intuitively, the rasterization pipeline of 3DGS consists of two sequential physical operations: mapping a 3D Gaussian onto the 2D space, and alpha-blending it along the ray. The Hessian field/matrix effectively maps the rendering processes, assigning high weights to proximal, highly salient primitives, and low weights to distant or occluded ones. Inspired by this forward rendering mechanism, we approximately decompose $\frac{{\partial  R^{\mathbf{v}}}}{{\partial G_i^k}}$ into two corresponding independent components: view-dependent visibility $V_i^{\mathbf{v}}$ and geometric projection $P_i^{\mathbf{v}}$:
\begin{equation}
\frac{{\partial  R^{\mathbf{v}}}}{{\partial G_i^k}} \approx{V_i^{\mathbf{v}}} \cdot {P_i^{\mathbf{v}}} ,
  \label{eq:9}
\end{equation}
where $V_i^{\mathbf{v}}$ describes how the attributes are influenced by the alpha-blending; and $P_i^{\mathbf{v}}$ defines how the attributes are effected by the projection from $\mathcal{M}$ to $\mathcal{I}$. In the following, we will explicitly model these two items. 

\vspace{0.5em}
\noindent\textbf{Modeling of Visibility.}
Rasterization in 3DGS is a view-dependent, sort-based blending process \cite{ali2024trimming,girish2024eagles,kwak2025modec}. The final color of each pixel $p$ is derived by alpha-blending the ordered set of overlapping Gaussians $\mathcal{N}$:
\begin{equation}
C(p) = \sum\limits_{i \in {\cal N}} \left [ {{{\tilde c}_i}} \cdot {\tilde \alpha _i}(p) \cdot T_i(p) \right ],
  \label{eq:11}
\end{equation}
where $\tilde{\alpha}_i(p)$ is the evaluated density at pixel $p$ multiplied by the standalone opacity $\alpha_i$, and $T_i(p) = \prod_{j=1}^{i-1} (1 - \tilde{\alpha}_j(p))$ denotes the accumulated transmittance. When the $i$-th Gaussian primitive is pruned, its gradient contribution to the final image quality is directly limited by $T_i$ and $\alpha_i$. Ignoring second-order occlusion differentials, the visibility $V_i^{\mathbf{v}}$ can be modeled as:
\begin{equation}
V_{i}^{\mathbf{v}} \approx \alpha_i \cdot T_i(p)  .
  \label{eq:12}
\end{equation}
However, calculating the exact accumulated transmittance requires complex depth sorting and ray marching, which is equivalent to a forward rendering pass. So we introduce a conservative zero-occlusion approximation (i.e., assuming $T_i \approx 1.0$). This models the worst-case scenario where the primitive is completely unoccluded, allowing its standalone opacity $\alpha_i$ to exclusively govern its visibility score. \par
\vspace{0.5em}
\noindent \textbf{Modeling of Geometric Projection $P_i^{\mathbf{v}}$.}
Beyond view-dependent visibility, the geometric mapping transformation changes the visual impact of the primitive. To quantify this influence, we formulate its perspective projection onto the 2D image space $\mathcal{I}$. Define the projection function $\Pi (\cdot): \mathbb{R}^3 \to \mathbb{R}^2$ as \cite{lee2024compact,lee2025optimized,li20253dhgs}:
\begin{equation}
\Pi(x_{i}^{\mathbf{v}}, y_{i}^{\mathbf{v}}, z_{i}^{\mathbf{v}}) = \left( f_x^{\mathbf{v}} \frac{x_{i}^{\mathbf{v}}}{z_{i}^{\mathbf{v}}} + u_x^{\mathbf{v}}, f_y^{\mathbf{v}} \frac{y_{i}^{\mathbf{v}}}{z_{i}^{\mathbf{v}}} + u_y^{\mathbf{v}} \right) ,
\label{eq:13}
\end{equation}
where $f^{\mathbf{v}}$ represents the focal length of the camera at viewpoint $\mathbf{v}$, which is specifically parameterized as $f_x^{\mathbf{v}}$ and $f_y^{\mathbf{v}}$ along the $x$ and $y$ axes of the image plane, respectively; The tuple $(u_x^{\mathbf{v}}, u_y^{\mathbf{v}})$ represents the principal point coordinates; Additionally, the vector $(x_{i}^{\mathbf{v}}, y_{i}^{\mathbf{v}}, z_{i}^{\mathbf{v}})^\top$ denotes the 3D center coordinate $\mu_i$ of the $i$-th Gaussian primitive, transformed into the local camera coordinate system of viewpoint $\mathbf{v}$.

Following the local affine approximation introduced in Elliptical Weighted Average (EWA) volume splatting \cite{zwicker2002ewa}, the perspective projection can be locally linearized using its Jacobian matrix $\mathbf{J}_{\Pi} \in \mathbb{R}^{2 \times 3}$. The image space visual displacement, also induced by a 3D spatial perturbation, is computed by this Jacobian. Consequently, the projection effect $P_i^{\mathbf{v}}$ is proportional to the energy of the projection matrix, which can be measured by
\begin{equation}
P_i^{\mathbf{v}} \approx \|\mathbf{J}_{\Pi}\|_F^2 = \text{Tr}(\mathbf{J}_{\Pi}^\top \mathbf{J}_{\Pi}) ,
\label{eq:14}
\end{equation}
where $\text{Tr}(\cdot)$ denotes the trace operator. Assuming a scaled orthographic projection as a local approximation of the perspective projection—which effectively ignores the eccentric view-dependent terms (i.e., assuming $x_i^{\mathbf{v}} \approx y_i^{\mathbf{v}} \approx 0$ locally) and assumes an isotropic focal length $f_x^{\mathbf{v}} \approx f_y^{\mathbf{v}} \approx f^{\mathbf{v}}$—the Jacobian energy simplifies to:
\begin{equation}
\text{Tr}(\mathbf{J}_{\Pi}^\top \mathbf{J}_{\Pi}) \approx 2\left(\frac{f^{\mathbf{v}}}{z^{\mathbf{v}}_i}\right)^2 .
\label{eq:16}
\end{equation}
Omitting the constant factor $2(f^{\mathbf{v}})^2$, the geometric influence fundamentally scales with the inverse square of the depth. To ensure numerical stability and prevent singularities when a primitive is exceptionally close to the camera plane, we introduce a stability constant $\eta=0.05$, yielding:
\begin{equation}
P_i^\mathbf{v} \propto \frac{1}{\left ( z_{i}^\mathbf{v} \right )^2+\eta} .
\label{eq:17}
\end{equation}
where $z_{i}^\mathbf{v}$ is the depth of the primitive relative to camera $\mathbf{v}$. Physically, Eq. (\ref{eq:17}) demonstrates that the importance of a primitive is highly depth-dependent: primitives closer to the camera induce a larger visual impact when pruned, whereas distant ones exert smaller influence on the image space.

By substituting Eqs. (\ref{eq:12}) and (\ref{eq:17}) into Eq. (\ref{eq:9}), and introducing the hyperparameter $\lambda^k$ to absorb the omitted constants and calibrate the varying sensitivities of attributes, we have
\begin{equation}
\frac{{\partial R^{\mathbf{v}}}}{{\partial \tilde{G}_i^k}} \approx \lambda^k \cdot \frac{\alpha_i}{(z_i^{\mathbf{v}})^2 + \eta} .
\label{eq:new1}
\end{equation}

\vspace{0.5em}
\noindent \textbf{Content-Adaptive Hyperparameters $\{\lambda^k\}$.}
The relative sensitivities of attributes vary between scenes. For example, opacity is critical for semi-transparent elements (e.g., foliage), while geometry dominates in rigid structures. A static $\lambda^k$ is hard to capture this variation. Therefore, we propose a content-adaptive mechanism to dynamically determine attribute sensitivity $\lambda^k$.

Specifically, we extract three statistical features from the Gaussian parameters to characterize the scene content: color variance $F_{col}$, opacity ambiguity $F_{opa}$, and scale anisotropy $F_{gem}$:
\begin{equation}
\begin{aligned}
F_{col} = \frac{1}{N}\sum_{i=1}^N (Y_i - \bar{Y})^2,
F_{opa} = \frac{1}{N}\sum_{i=1}^N \alpha_i (1 - \alpha_i),
F_{gem} = \frac{1}{N}\sum_{i=1}^N \frac{\max(s_i)}{\min(s_i) + \eta}.
\end{aligned}
\label{eq:features}
\end{equation}
where $Y_i$ is the perceptual luma computed by $c_i$, and $\bar{Y}$ is the mean luma across the scene. Physically, $F_{col}$ quantifies the richness of high-frequency textures; $F_{opa}$ is a concave function to evaluate the proportion of semi-transparent regions; and $F_{gem}$ measures the average structural stretch of the primitives. And the content-adaptive hyperparameter $\lambda^k$ is obtained: 
\begin{equation}
{\lambda ^k} = \frac{{{{\tilde \lambda }^k}}}{{\sum\limits_{j \in \{ gem,col,opa\} } {{{\tilde \lambda }^j}} }},
\label{eq:19}
\end{equation}
where the unnormalized $\tilde{\lambda}^k$ are calculated as:
\begin{equation}
\begin{aligned}
\tilde{\lambda}^{gem} = \frac{\ln(F_{gem} + \eta)}{\mathbb{E}[\ln(F_{gem})]}, \quad
\tilde{\lambda}^{col} = \frac{F_{col}}{\mathbb{E}[F_{col}]}, \quad
\tilde{\lambda}^{opa} = \frac{F_{opa}}{\mathbb{E}[F_{opa}]}.
\end{aligned}
\label{eq:18}
\end{equation}
We calibrate these features using their expectations $\mathbb{E}[\cdot]$ to balance their magnitude discrepancies, applying logarithmic smoothing to $F_{gem}$ to suppress extreme stretches. 

Finally, the rendering-aware Hessian weight $w_i^k$ in Eq. (\ref{eq:8}) is derived as: 
\begin{equation}
w_i^k = {\lambda ^k} \cdot \frac{1}{|\mathcal{V}|} \sum\limits_{{\mathbf{v}} \in \mathcal{V}} \left[ \frac{\alpha_i}{\left ( z_{i}^{\mathbf{v}} \right )^2 + \eta} \right] .
  \label{eq:20}
\end{equation}

\subsection{Super-efficient Pruning Process}
\label{sec 3.3}
During the pruning execution, for a pre-trained 3DGS scene containing $N$ primitives, we couple the estimated Hessian field with parameter values following our prior derivations to calculate the final rendering-free importance score for each individual Gaussian primitive. Subsequently, all primitives are globally ranked based on these evaluated scores. Given a target pruning ratio $\beta \in (0, 1)$, we permanently remove the bottom $\beta \cdot N$ primitives associated with the lowest importance scores, as shown in Fig. \ref{fig2}. By entirely bypassing actual rendering passes, our method achieves a highly efficient pruning process. 


\section{Experiments}
\subsection{Experiment Settings}
\textbf{Datasets.} We evaluated our method on the same challenging real-world scenes as 3D-GS. We used all nine scenes from the Mip-NeRF 360 dataset\cite{barron2022mipnerf360}, which contains five outdoor and four indoor scenes, each featuring complex central objects or viewing areas and detailed backgrounds. Additionally, two outdoor scenes, $truck$ and $train$, were taken from the Tanks \& Temples dataset\cite{knapitsch2017tanks}, and two indoor scenes, $drjohnson$ and $playroom$, were taken from the Deep Blending dataset\cite{hedman2018deep}. For consistency, we used the COLMAP camera pose estimates provided in the original 3D-GS creator's pre-experiments\cite{kerbl20233d}.  

\vspace{0.5em}
\noindent\textbf{Implementation Details.}
Our REFINE is a plug-and-play, purely post-proces- sing pruning pipeline that can be seamlessly applied to any pre-trained 3DGS model \cite{kerbl20233d}. To highlight the effectiveness of the evaluation metric itself, all comparisons regarding pruning effects were conducted under a \textit{zero-shot} condition. That is, after removing primitives based on the algorithm, strictly no subsequent fine-tuning rendering optimization was performed.  

\vspace{0.5em}
\noindent\textbf{Baseline Methods.}  We compared our REFINE with four representative and state-of-the-art 3DGS pruning methods, including 
GHAP\cite{wang2025gaussian}, LightGaussian\cite{fan2024lightgaussian}, MesonGS\cite{xie2024mesongs}, and PUP 3D-GS\cite{hanson2025pup}.\par

\subsection{Experimental Results} \label{sec4.2}

We comprehensively tested the performance of pruning various methods in terms of rendering fidelity and computational complexity across pruning ratios from 10\% to 70\%.  

\begin{table}[t]
\centering
\setlength{\tabcolsep}{3pt}
\renewcommand{\arraystretch}{1.0}
\definecolor{bestred}{HTML}{FF0000}
\definecolor{secondblue}{HTML}{3531FF}
\caption{Quantitative Comparison of Different Pruning Methods. The best and second-best are highlighted in \textcolor{bestred}{red} and \textcolor{secondblue}{blue}, respectively.}
\vspace{-0.25cm}
\begin{adjustbox}{max width=\linewidth}
\begin{tabular}{c|c|ccc|ccc|ccc|ccc}
\toprule
& & \multicolumn{3}{c|}{\textbf{Ratio = 0.1}}
& \multicolumn{3}{c|}{\textbf{Ratio = 0.3}}
& \multicolumn{3}{c|}{\textbf{Ratio = 0.5}}
& \multicolumn{3}{c}{\textbf{Ratio = 0.7}} \\
\multirow{-2}{*}{\textbf{Dataset}}
& \multirow{-2}{*}{\textbf{Method}}
& PSNR$\uparrow$ & SSIM$\uparrow$ & LPIPS$\downarrow$
& PSNR$\uparrow$ & SSIM$\uparrow$ & LPIPS$\downarrow$
& PSNR$\uparrow$ & SSIM$\uparrow$ & LPIPS$\downarrow$
& PSNR$\uparrow$ & SSIM$\uparrow$ & LPIPS$\downarrow$ \\ \hline

& original 3D GS \cite{kerbl20233d}
& 27.35 & 0.814 & 0.217
& 27.35 & 0.814 & 0.217
& 27.35 & 0.814 & 0.217
& 27.35 & 0.814 & 0.217 \\

& GHAP \cite{wang2025gaussian}
& 20.65 & {\color[HTML]{3531FF} 0.607} & {\color[HTML]{3531FF} 0.445}
& 20.62 & 0.594 & 0.455
& 20.44 & 0.571 & 0.470
& 19.93 & 0.530 & 0.494 \\

& \CZADD{LightGaussian} \cite{fan2024lightgaussian}
& {\color[HTML]{FF0000} 27.34} & {\color[HTML]{FF0000} 0.814} & {\color[HTML]{FF0000} 0.217}
& {\color[HTML]{FF0000} 27.33} & {\color[HTML]{FF0000} 0.814} & {\color[HTML]{FF0000} 0.217}
& {\color[HTML]{3531FF} 26.92} & {\color[HTML]{3531FF} 0.806} & {\color[HTML]{3531FF} 0.225}
& 24.73 & 0.762 & 0.264 \\

& MesonGS \cite{xie2024mesongs}
& {\color[HTML]{3531FF} {27.33}}
& {\color[HTML]{FF0000} {0.814}}
& {\color[HTML]{FF0000} {0.217}}
& {\color[HTML]{FF0000} {27.33}}
& {\color[HTML]{FF0000} {0.814}}
& {\color[HTML]{FF0000} {0.217}}
& {\color[HTML]{FF0000} {27.06}}
& {\color[HTML]{FF0000} {0.809}}
& {\color[HTML]{FF0000} {0.222}}
& {\color[HTML]{FF0000} {25.46}}
& {\color[HTML]{3531FF} {0.783}}
& {\color[HTML]{FF0000} {0.245}} \\

& PUP \cite{hanson2025pup}
& {\color[HTML]{FF0000} 27.34} & {\color[HTML]{FF0000} 0.814} & {\color[HTML]{FF0000} 0.217}
& 27.24 & 0.812 & 0.219
& 26.07 & 0.790 & 0.241
& {\color[HTML]{3531FF} 25.12} & {\color[HTML]{FF0000} 0.786} & {\color[HTML]{3531FF} 0.259} \\

\multirow{-5}{*}{\rotatebox{90}{\makecell{MipNeRF \\ 360}}}
& REFINE (\textbf{Ours})
& {\color[HTML]{FF0000} 27.34} & {\color[HTML]{FF0000} 0.814} & {\color[HTML]{FF0000} 0.217}
& {\color[HTML]{3531FF} 27.29} & {\color[HTML]{3531FF} 0.813} & {\color[HTML]{3531FF} 0.218}
& 26.61 & 0.800 & 0.233
& 24.43 & 0.745 & 0.285 \\ \hline

& original 3D GS \cite{kerbl20233d}
& 23.39 & 0.842 & 0.183
& 23.39 & 0.842 & 0.183
& 23.39 & 0.842 & 0.183
& 23.39 & 0.842 & 0.183 \\

& GHAP \cite{wang2025gaussian}
& 16.84 & 0.592 & 0.467
& 17.00 & 0.587 & 0.473
& 16.97 & 0.576 & 0.482
& 16.82 & 0.553 & 0.501 \\

& \CZADD{LightGaussian} \cite{fan2024lightgaussian}
& {\color[HTML]{FF0000} 23.39} & {\color[HTML]{FF0000} 0.842} & {\color[HTML]{FF0000} 0.183}
& {\color[HTML]{FF0000} 23.38} & {\color[HTML]{FF0000} 0.841} & {\color[HTML]{FF0000} 0.183}
& {\color[HTML]{FF0000} 23.24} & {\color[HTML]{FF0000} 0.835} & {\color[HTML]{FF0000} 0.190}
& {\color[HTML]{3531FF} 21.96} & {\color[HTML]{3531FF} 0.797} & {\color[HTML]{3531FF} 0.225} \\

& MesonGS \cite{xie2024mesongs}
& {\color[HTML]{3531FF} {23.38}}
& {\color[HTML]{3531FF} {0.841}}
& {\color[HTML]{3531FF} {0.184}}
& {\color[HTML]{FF0000} {23.38}}
& {\color[HTML]{FF0000} {0.841}}
& {\color[HTML]{3531FF} {0.184}}
& {\color[HTML]{3531FF} {23.18}}
& {\color[HTML]{3531FF} {0.834}}
& {\color[HTML]{3531FF} {0.191}}
& {\color[HTML]{FF0000} {22.04}}
& {\color[HTML]{FF0000} {0.803}}
& {\color[HTML]{FF0000} {0.220}} \\

& PUP \cite{hanson2025pup}
& {\color[HTML]{FF0000} 23.39} & {\color[HTML]{FF0000} 0.842} & {\color[HTML]{FF0000} 0.183}
& {\color[HTML]{FF0000} 23.38} & {\color[HTML]{FF0000} 0.841} & {\color[HTML]{3531FF} 0.184}
& {\color[HTML]{3531FF} 23.18} & 0.832 & 0.193
& 21.39 & 0.787 & 0.233 \\

\multirow{-5}{*}{\rotatebox{90}{\makecell{Tanks \& \\ Temples}}}
& REFINE (\textbf{Ours})
& {\color[HTML]{3531FF} 23.38} & {\color[HTML]{3531FF} 0.841} & {\color[HTML]{FF0000} 0.183}
& {\color[HTML]{3531FF} 23.33} & {\color[HTML]{3531FF} 0.839} & 0.185
& 22.97 & 0.828 & 0.196
& 20.98 & 0.776 & 0.243 \\ \hline

& original 3D GS \cite{kerbl20233d}
& 29.52 & 0.903 & 0.242
& 29.52 & 0.903 & 0.242
& 29.52 & 0.903 & 0.242
& 29.52 & 0.903 & 0.242 \\

& GHAP \cite{wang2025gaussian}
& 23.12 & {\color[HTML]{3531FF} 0.768} & {\color[HTML]{3531FF} 0.456}
& 23.18 & {\color[HTML]{3531FF} 0.767} & 0.459
& 23.16 & 0.763 & 0.464
& 22.92 & 0.752 & 0.475 \\

& \CZADD{LightGaussian} \cite{fan2024lightgaussian}
& {\color[HTML]{3531FF} 29.52} & {\color[HTML]{FF0000} 0.903} & {\color[HTML]{FF0000} 0.242}
& {\color[HTML]{3531FF} 29.51} & {\color[HTML]{FF0000} 0.903} & {\color[HTML]{3531FF} 0.243}
& {\color[HTML]{3531FF} 29.36} & {\color[HTML]{3531FF} 0.900} & {\color[HTML]{3531FF} 0.247}
& 28.38 & 0.883 & 0.269 \\

& MesonGS \cite{xie2024mesongs}
& {\color[HTML]{FF0000} {29.53}}
& {\color[HTML]{FF0000} {0.903}}
& {\color[HTML]{FF0000} {0.242}}
& {\color[HTML]{FF0000} {29.53}}
& {\color[HTML]{FF0000} {0.903}}
& {\color[HTML]{FF0000} {0.242}}
& {\color[HTML]{FF0000} {29.43}}
& {\color[HTML]{FF0000} {0.902}}
& {\color[HTML]{FF0000} {0.245}}
& {\color[HTML]{3531FF} {28.68}}
& {\color[HTML]{3531FF} {0.890}}
& {\color[HTML]{3531FF} {0.262}} \\

& PUP \cite{hanson2025pup}
& {\color[HTML]{3531FF} 29.52} & {\color[HTML]{FF0000} 0.903} & {\color[HTML]{FF0000} 0.242}
& {\color[HTML]{3531FF} 29.51} & {\color[HTML]{FF0000} 0.903} & {\color[HTML]{3531FF} 0.243}
& 29.28 & 0.899 & 0.248
& {\color[HTML]{FF0000} 29.14} & {\color[HTML]{FF0000} 0.894} & {\color[HTML]{FF0000} 0.256} \\

\multirow{-5}{*}{\rotatebox{90}{\makecell{Deep \\ Blending}}}
& REFINE (\textbf{Ours})
& {\color[HTML]{3531FF} 29.52} & {\color[HTML]{FF0000} 0.903} & {\color[HTML]{FF0000} 0.242}
& 29.49 & {\color[HTML]{FF0000} 0.903} & {\color[HTML]{3531FF} 0.243}
& 29.28 & 0.899 & 0.249
& 27.93 & 0.871 & 0.282 \\

\bottomrule
\end{tabular}
\end{adjustbox}
\label{tab:comp_quality}
\end{table}

\vspace{0.5em}
\noindent \textbf{Comparisons of Rendering Fidelity.} We adopted three widely accepted image quality evaluation metrics: Peak Signal-to-Noise Ratio (PSNR, higher is better), Structural Similarity Index (SSIM, higher is better), and Learned Perceptual Image Patch Similarity (LPIPS, lower is better) \cite{zhang2018unreasonable}. From Table \ref{tab:comp_quality}, among the methods evaluated, there are significant differences in performance. \CZADD{LightGaussian demonstrates strong rendering fidelity across different pruning ratios; for instance, on the MipNeRF 360 dataset, it maintains a PSNR of 26.92 even when the pruning ratio reaches 50\%.} Meanwhile, GHAP yields a lower PSNR without fine-tuning, as its optimal transport formulation is designed for iterative cluster reconstruction.

Across the 10\% to 70\% pruning, our REFINE demonstrates exceptional efficacy. At a 10\% removal ratio on MipNeRF 360, REFINE achieves a PSNR of 27.34, an SSIM of 0.814, and an LPIPS of 0.217, performing fully on par with \CZADD{LightGaussian and} the state-of-the-art rendering-based method, PUP. Notably, on the Tanks \& Temples dataset, REFINE remains highly competitive with \CZADD{strong baselines} across all pruning ratios (e.g., \CZADD{22.97 vs. 23.24 for LightGaussian} at 50\% pruning). Even when pushed to a 50\% pruning ratio on MipNeRF 360, REFINE maintains strong visual fidelity (\CZADD{26.61 vs. 26.92 for LightGaussian and 27.06 for MesonGS}). To achieve extreme rendering-free speed, REFINE explicitly approximates and omits second-order coupling effects. However, we note a performance drop at an extreme pruning ratio (Ratio = 0.7) on the Deep Blending dataset (e.g., PSNR drops to 27.93). This behavior is physically intuitive. Deep Blending scenes are characterized by extremely dense primitive distributions and severe occlusions. At extreme pruning ratios, the cross-primitive coupling effect, which is intentionally omitted in our rendering-free analytical formulation for the sake of super-efficiency, becomes non-negligible. \par

\begin{table}[t]
\centering
\setlength{\tabcolsep}{3pt} 
\renewcommand{\arraystretch}{1.0} 
\definecolor{bestred}{HTML}{FF0000}
\definecolor{secondblue}{HTML}{3531FF}
\caption{Computational Efficiency Comparison of Different Pruning Methods. The best and second-best are highlighted in \textcolor{bestred}{red} and \textcolor{secondblue}{blue}, respectively.}
\vspace{-0.25cm}
\begin{adjustbox}{max width=\linewidth}
\begin{tabular}{c|c|cc|cc|cc|cc}
\toprule
                                   &                                   & \multicolumn{2}{c|}{\textbf{Ratio = 0.1}}                    & \multicolumn{2}{c|}{\textbf{Ratio = 0.3}}                    & \multicolumn{2}{c|}{\textbf{Ratio = 0.5}}                    & \multicolumn{2}{c}{\textbf{Ratio = 0.7}}                    \\
\multirow{-2}{*}{\textbf{Dataset}} & \multirow{-2}{*}{\textbf{Method}} & Time (s) $\downarrow$ & GFLOPs$\downarrow$ & Time (s)$\downarrow$ & GFLOPs$\downarrow$ & Time (s)$\downarrow$ & GFLOPs$\downarrow$ & Time (s)$\downarrow$ & GFLOPs$\downarrow$ \\ \hline

                                   & GHAP \cite{wang2025gaussian}
                                   & {\color[HTML]{3531FF} 14.56} & {\color[HTML]{3531FF} 244.30}
                                   & {\color[HTML]{3531FF} 12.42} & {\color[HTML]{3531FF} 190.26}
                                   & {\color[HTML]{3531FF} 10.70} & {\color[HTML]{3531FF} 136.23}
                                   & {\color[HTML]{3531FF} 8.96}  & {\color[HTML]{3531FF} 82.19} \\

                                   & \CZADD{LightGaussian} \cite{fan2024lightgaussian}
                                   & {42.14} & { 2855.78}
                                   & {37.60} & { 2855.78}
                                   & {30.07} & { 2855.78}
                                   & {26.88} & { 2855.78} \\

                                   & MesonGS \cite{xie2024mesongs}
                                   & 48.99 & 9539.19
                                   & 44.80 & 9539.19
                                   & 41.43 & 9539.19
                                   & 33.70 & 9539.19 \\

                                   & PUP \cite{hanson2025pup}
                                   & 44.01 & 9582.81
                                   & 40.28 & 9582.81
                                   & 37.02 & 9582.81
                                   & 38.15 & 9582.81 \\

\multirow{-5}{*}{\rotatebox{90}{\makecell{MipNeRF \\ 360}}}
                                   & REFINE (\textbf{Ours})
                                   & {\color[HTML]{FF0000} 5.90} & {\color[HTML]{FF0000} 3.14}
                                   & {\color[HTML]{FF0000} 5.46} & {\color[HTML]{FF0000} 3.14}
                                   & {\color[HTML]{FF0000} 3.85} & {\color[HTML]{FF0000} 3.14}
                                   & {\color[HTML]{FF0000} 2.55} & {\color[HTML]{FF0000} 3.14} \\ \hline

                                   & GHAP \cite{wang2025gaussian}
                                   & {\color[HTML]{3531FF} 7.67} & {\color[HTML]{3531FF} 129.59}
                                   & {\color[HTML]{3531FF} 6.62} & {\color[HTML]{3531FF} 100.93}
                                   & {\color[HTML]{3531FF} 5.71} & {\color[HTML]{3531FF} 72.26}
                                   & {\color[HTML]{3531FF} 4.84} & {\color[HTML]{3531FF} 43.60} \\

                                   & \CZADD{LightGaussian} \cite{fan2024lightgaussian}
                                   & {28.88} & 2128.25
                                   & {26.62} & 2128.25
                                   & {24.70} & 2128.25
                                   & {22.64} & 2128.25 \\

                                   & MesonGS \cite{xie2024mesongs}
                                   & 21.99 & 7104.05
                                   & 20.27 & 7104.05
                                   & 18.47 & 7104.05
                                   & 14.11 & 7104.05 \\

                                   & PUP \cite{hanson2025pup}
                                   & 18.90 & 7136.18
                                   & 17.75 & 7136.18
                                   & 15.92 & 7136.18
                                   & 16.75 & 7136.18 \\

\multirow{-5}{*}{\rotatebox{90}{\makecell{Tanks \& \\ Temples}}}
                                   & REFINE (\textbf{Ours})
                                   & {\color[HTML]{FF0000} 3.34} & {\color[HTML]{FF0000} 1.67}
                                   & {\color[HTML]{FF0000} 2.70} & {\color[HTML]{FF0000} 1.67}
                                   & {\color[HTML]{FF0000} 1.99} & {\color[HTML]{FF0000} 1.67}
                                   & {\color[HTML]{FF0000} 1.36} & {\color[HTML]{FF0000} 1.67} \\ \hline

                                   & GHAP \cite{wang2025gaussian}
                                   & {\color[HTML]{3531FF} 13.04} & {\color[HTML]{3531FF} 216.19}
                                   & {\color[HTML]{3531FF} 10.93} & {\color[HTML]{3531FF} 168.37}
                                   & {\color[HTML]{3531FF} 9.37}  & {\color[HTML]{3531FF} 120.55}
                                   & {\color[HTML]{3531FF} 7.88}  & {\color[HTML]{3531FF} 72.74}  \\

                                   & \CZADD{LightGaussian} \cite{fan2024lightgaussian}
                                   & {33.69} & 3301.74
                                   & {32.39} & 3301.74
                                   & {29.16} & 3301.74
                                   & {25.96} & 3301.74 \\

                                   & MesonGS \cite{xie2024mesongs}
                                   & 24.41 & 11017.66
                                   & 21.09 & 11017.66
                                   & 18.00 & 11017.66
                                   & 12.83 & 11017.66 \\

                                   & PUP \cite{hanson2025pup}
                                   & 19.94 & 11067.59
                                   & 16.92 & 11067.59
                                   & 13.84 & 11067.59
                                   & 15.15 & 11067.59 \\

\multirow{-5}{*}{\rotatebox{90}{\makecell{Deep \\ Blending}}}
                                   & REFINE (\textbf{Ours})
                                   & {\color[HTML]{FF0000} 5.37} & {\color[HTML]{FF0000} 1.77}
                                   & {\color[HTML]{FF0000} 4.31} & {\color[HTML]{FF0000} 1.77}
                                   & {\color[HTML]{FF0000} 3.33} & {\color[HTML]{FF0000} 1.77}
                                   & {\color[HTML]{FF0000} 1.42} & {\color[HTML]{FF0000} 1.77} \\

\bottomrule
\end{tabular}
\end{adjustbox}
\label{tab::efficiency}
\end{table}

\begin{figure}[]
  \centering
  \includegraphics[height=15.cm]{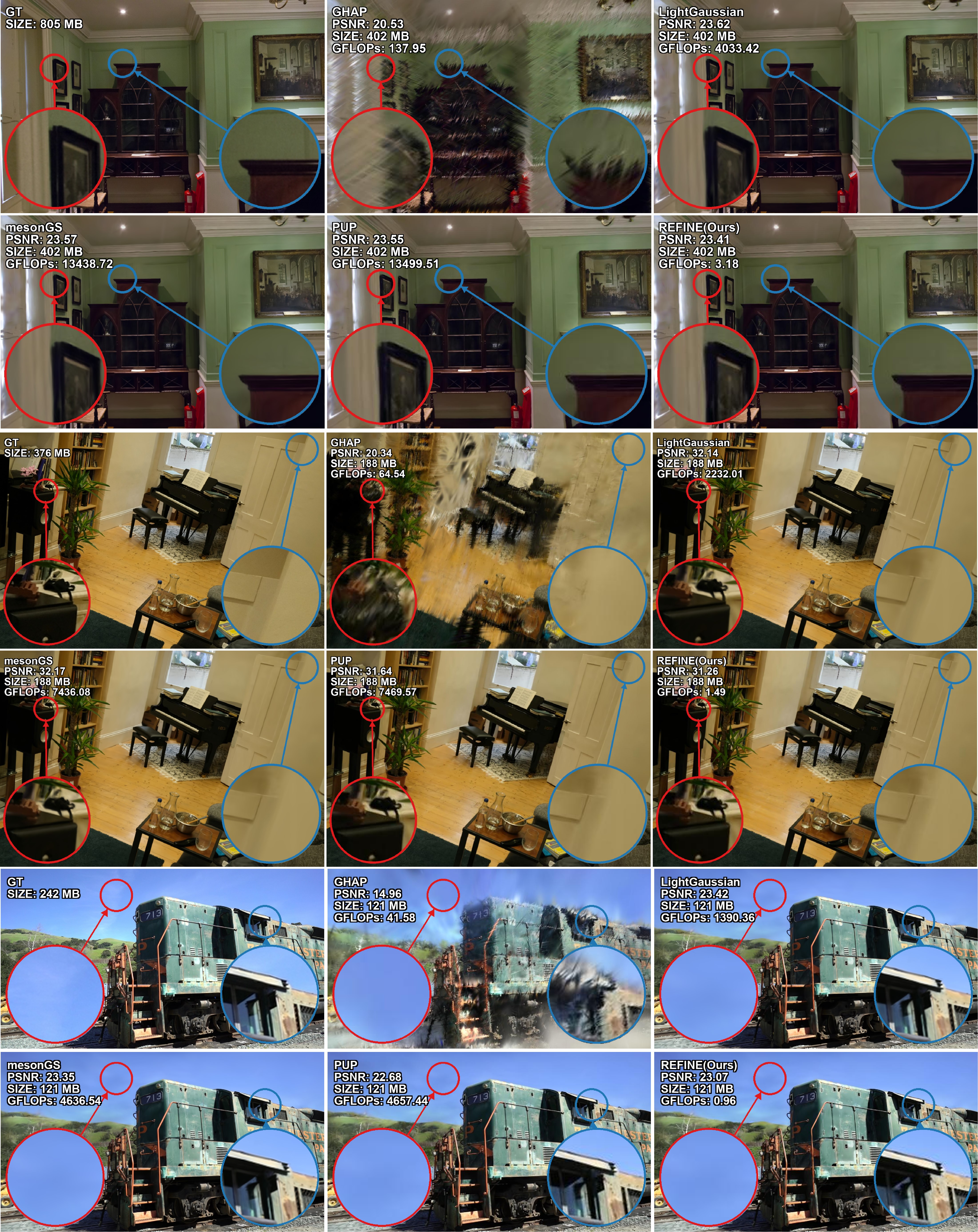}
  \caption{ \CZADD{ Visual comparison after 50\% pruning using REFINE and other methods. Top: $drjohnson$ from Deep Blending. Middle: $room$ from Mip-NeRF 360. Bottom: $train$ from Tanks \& Temples. Zooming in for details.}
  }
  \label{fig:visual}
\end{figure}

\vspace{0.5em}
\noindent\textbf{Comparisons of Computational Complexity}. 
Computational complexity is the key to applying pruning algorithms in practice. We recorded the total processing time (in seconds) and the total computational cost (GFLOPs) required to evaluate primitive importance for each method \cite{qiao2026meaa}\cite{weloday2026lwmscnn}. As reported in Table \ref{tab::efficiency}, rendering-based methods (e.g., PUP and MesonGS) incur massive computational burdens, demanding up to $\sim$11,000 GFLOPs to evaluate a single scene. 
On the contrary, REFINE's computational cost remains remarkably low, requiring merely 3.14, 1.67, and 1.77 GFLOPs on the MipNeRF 360, Tanks \& Temples, and Deep Blending datasets, respectively. This represents a staggering reduction in computational cost by over $3000\times$ compared to PUP. Correspondingly, REFINE achieves the fastest processing speed across all scenarios, completing the entire pruning execution in as little as 1.36 seconds. This efficiency stems from our closed-form solution Eq. (\ref{eq:20}). By bypassing rendering entirely, REFINE relies solely on $\mathcal{O}(N)$ parameter space algebraic operations. Consequently, it serves as an super-efficient, plug-and-play module ideal for resource-constrained devices, visual comparison as shown in Fig. \ref{fig:visual}. Note that while GFLOPs are reduced by $>3000\times$, the latency reduction is relatively smaller ($<20\times$). This is because rendering-based methods fully saturate the highly parallelized GPU during rasterization, masking their massive computational burden, whereas REFINE achieves its speed with minimal hardware utilization. Furthermore, the total latency includes fixed I/O overhead (e.g., sorting and saving primitives). A detailed FLOPs breakdown per primitive is provided in the Supplementary Material.
\par

\subsection{Ablation Study}
\label{sec5.4}
\begin{wrapfigure}{r}{0.50\textwidth} %
  \vspace{-40pt} 
  \centering
  \captionof{table}{Ablation study on attribute modulation at a 50\% pruning ratio.}
  \label{tab:ablation}
  \setlength{\tabcolsep}{2pt} 
  \renewcommand{\arraystretch}{1.0} 
  \small 
  \begin{tabular}{lccc}
    \toprule
    \textbf{Conf.} & PSNR $\uparrow$ & SSIM $\uparrow$ & LPIPS $\downarrow$ \\ 
    \midrule
    w/o $V_i^\mathbf{v}$  & 27.74 & 0.849 & 0.212 \\
    w/o $P_i^\mathbf{v}$  & 27.24 & 0.842 & 0.214 \\ 
    \midrule
    Equal $\lambda^k$      & 27.82 & 0.852 & 0.209 \\ 
    \midrule
    \textbf{Ours}          & \textbf{28.35} & \textbf{0.862} & \textbf{0.198} \\ 
    \bottomrule
  \end{tabular}
  
  \vspace{10pt} 
  \includegraphics[width=0.83\linewidth]{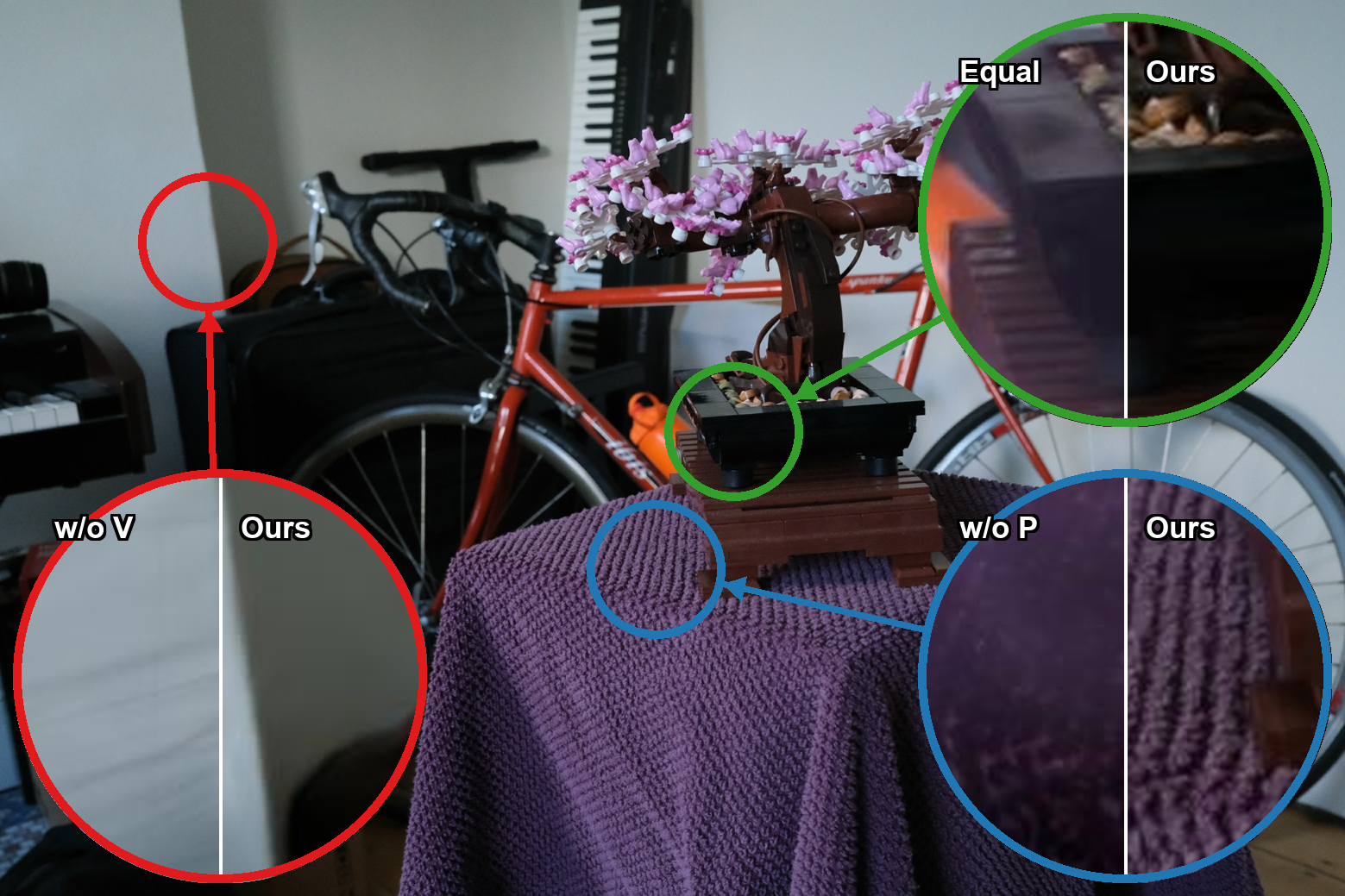}
  \captionof{figure}{Visual ablation study after 50\% pruning.}
  \label{fig:ablation}
  \vspace{-35pt} 
\end{wrapfigure}
To better understand our proposed REFINE method, we conducted thorough ablation studies at a 50\% pruning ratio. Table \ref{tab:ablation} reports the average performance across the evaluated datasets ($bicycle$, $bonsai$, and $kitchen$) for different attribute configurations. Fig. \ref{fig:ablation} presents a visual comparison of the ablation study.

\vspace{0.5em}
\noindent \textbf{Effectiveness of Components.} We first evaluated extreme configurations where primitive importance is dictated solely by a single component (e.g., \textit{w/o $V_i^\mathbf{v}$} relying exclusively on geometric projection, or \textit{w/o $P_i^\mathbf{v}$} relying purely on view-dependent visibility). Table \ref{tab:ablation} shows that single component strategies perform poorly (e.g., PSNR 27.24 for \textit{w/o $P_i^\mathbf{v}$}), confirming the necessity of jointly modeling all components.

\vspace{0.5em}
\noindent \textbf{Superiority of the Content Adaptive Hyperparameter.} A static weight distribution (\textit{Equal} $\lambda ^k$) yields a suboptimal mean PSNR of 27.82. By dynamically adjusting to scene statistics, our adaptive method (\textbf{Ours}) boosts this to 28.35. 

\vspace{0.5em}
\noindent \textbf{Validation of Structured Approximation.} 
We quantitatively verified the rationality of the two assumptions on primitive independence and attribute orthogonality in parameter space construction by analyzing the energy distribution of the Hessian matrix. \par

\textit{Verification of Primitive Independence}:
Although the 3DGS rendering process involves alpha blending induced occlusion coupling, we assume the Hessian matrix $\mathbf{H}$ has a significant block-diagonal structure. To verify this, we randomly selected 100 Gaussian primitives in evaluated datasets, used the $30$-Nearest Neighbors to sample high-density local regions to construct a coupling test set, and calculated the exact Hessian matrix regarding the rendering loss. We used the Diagonal Energy Ratio (DER) as a quantitative metric, defined as $\mathrm{DER} = \sum_{i=1}^N \|\mathbf{H}_{i,i}\|_F^2 \,/\, \|\mathbf{H}\|_F^2$, where $\mathbf{H}_{i,i}$ represents the autocorrelation block of the $i$-th Gaussian in the sampled Hessian matrix, and $\mathbf{H}$ represents the sampled Hessian matrix. A DER value closer to 1 indicates that energy is primarily distributed on the diagonal, implying lower coupling energy.\par

\begin{figure}[htbp]
  \centering
  \begin{subfigure}[b]{0.38\textwidth} 
    \centering
    \includegraphics[width=\linewidth]{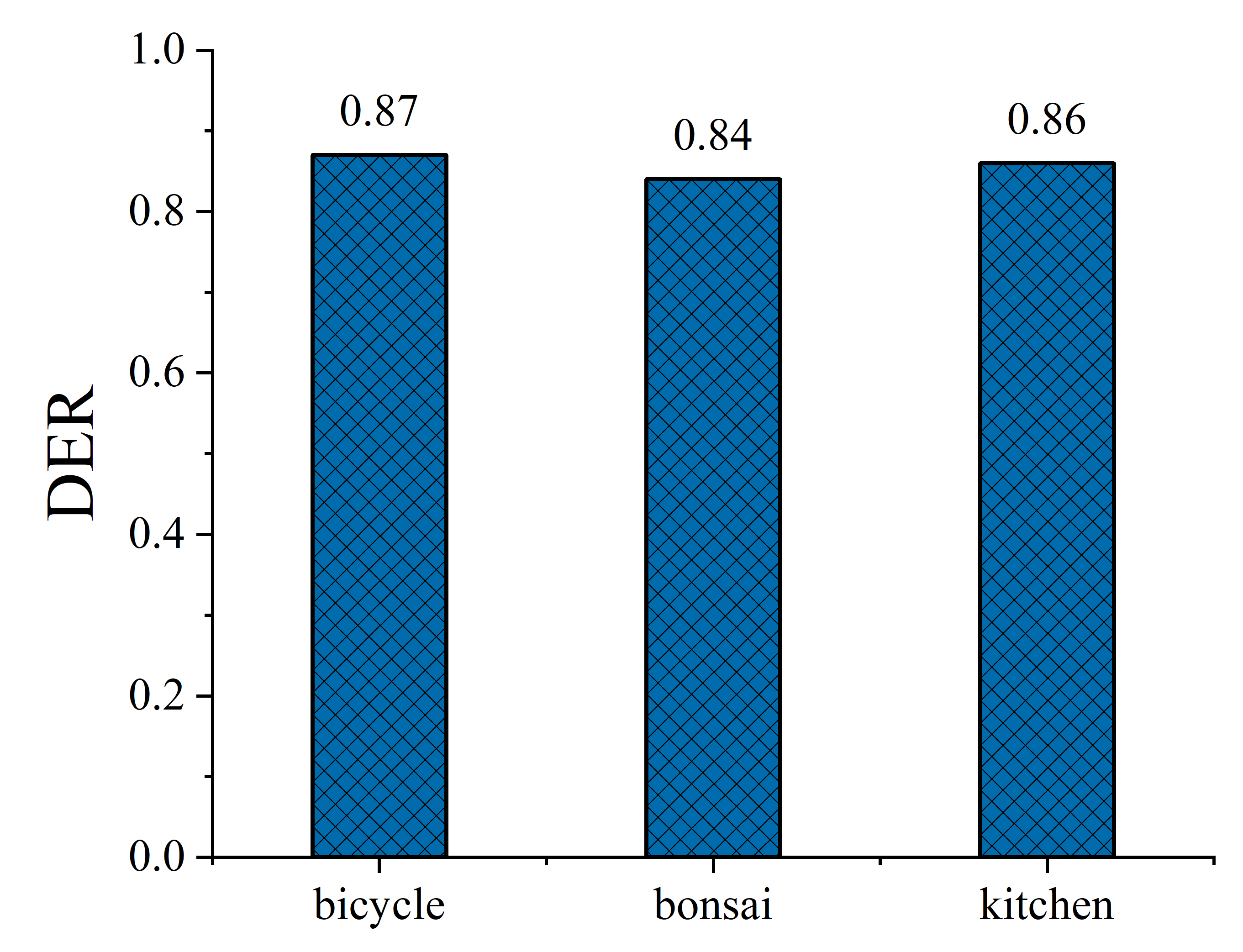}
    \caption{Average DER of Hessian matrices}
    \label{fig:sub_a}
  \end{subfigure}
  \hspace{0.08\textwidth} 
  \begin{subfigure}[b]{0.34\textwidth}
    \centering
    \includegraphics[width=\linewidth]{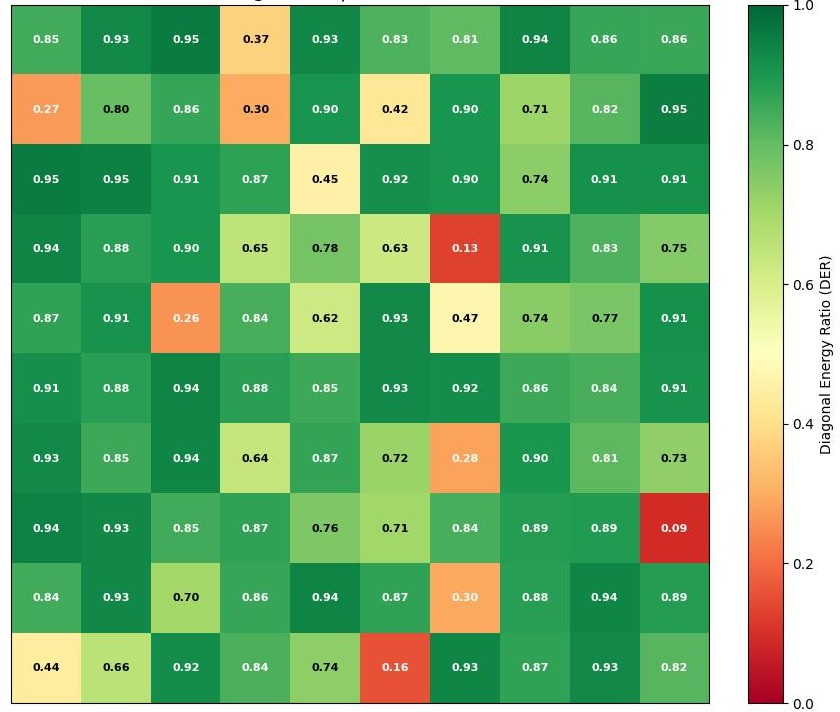}
    \caption{Details of DER on $bonsai$}
    \label{fig:sub_b}
  \end{subfigure}
  \caption{DER of Hessian matrices for different Gaussian primitives.}
  \label{fig:DER}
\end{figure}

Experimental results are shown in Fig. \ref{fig:DER} (a). Even in regions with dense spatial overlap, the DER remains as high as 86\%. This means that nearly 90\% of the energy is concentrated in the diagonal blocks, and gradient interactions between Gaussian primitives account for only 14\%.  The heatmap in Fig. \ref{fig:DER} (b) further intuitively displays the DER energy values of 100 randomly sampled Gaussian primitives on $bonsai$. This finding provides strong statistical support for ignoring the off-diagonal blocks.

\textit{Verification of  Attribute Orthogonality}.
Regarding the parameter space within a single Gaussian, we verified the orthogonality of its parameter subspaces. Combining the specific implementation process, we categorized attributes into Geometry, Color, and Opacity. We similarly used the DER as the quantitative metric. 
\begin{wraptable}{r}{0.45\textwidth}
  \vspace{-15pt} 
  \caption{Average DER of Hessian matrices across different scenes.}
  \label{tab_III}
  \centering
  \small 
  \begin{tabular}{cc} 
    \toprule
    Sequences & Average DER \\ 
    \midrule
    bicycle & 0.91 \\
    bonsai & 0.78 \\
    kitchen & 0.67 \\ 
    \midrule
    \textbf{Average} & \textbf{0.79} \\ 
    \bottomrule
  \end{tabular}
  \vspace{-15pt} 
\end{wraptable}
As shown in Table \ref{tab_III}, the average DER reaches 0.79, indicating that the majority of the second-order optimization energy is highly concentrated on the diagonal blocks. Specifically, the $bicycle$ scene exhibits the highest DER of 0.91, which perfectly aligns with our theoretical assumption that cross-primitive coupling effects can be safely decoupled. Notably, $kitchen$ scene exhibits a lower DER (0.67). This is primarily because $kitchen$ is a complex indoor environment characterized by dense, overlapping surfaces and severe occlusions. Consequently, a massive number of Gaussian primitives heavily overlap along the same rendering rays, leading to stronger gradient dependencies and a relatively lower diagonal energy concentration.\par

\section{Conclusion and Discussion}
In this paper, we presented REFINE, a super-efficient, rendering-free pruning framework for 3DGS. By analytically approximating a rendering-aware Hessian field to evaluate primitive importance, REFINE successfully breaks the trade-off between visual fidelity and computational cost, achieving unprecedented acceleration over existing rendering-based methods. 
Beyond post-training pruning, our method also offers promising avenues for broader 3DGS optimization. It can be seamlessly integrated into rate-distortion optimization to guide efficient compression, or serve as a dynamic regularizer to accelerate the standard 3DGS training process. 

Although our method achieves impressive performance,  our primitive independence assumption may lead to quality degradation at \textit{extreme} pruning ratios where cross-primitive coupling becomes severe. Future work could explore incorporating lightweight, localized re-rendering passes or low-rank off-diagonal approximations to mitigate this while preserving real-time efficiency.

\section*{Acknowledgements}
This work was supported in part by the National Natural Science Foundation of China under Grants 62422118, and in part by the Hong Kong Research Grants Council under Grants N\_CityU1114/25 and 11219324.

%
%
\bibliographystyle{splncs04}
\bibliography{main}

\newpage

\appendix
\setcounter{table}{0}
\setcounter{figure}{0}
\setcounter{equation}{0}

\renewcommand{\thetable}{B.\arabic{table}}
\renewcommand{\thefigure}{B.\arabic{figure}}
\renewcommand{\theequation}{B.\arabic{equation}}

\section{Supplementary Material} \label{sec:sup_Computation}

\noindent \textbf{GFLOPs vs. Latency \& Computation Methodology.}
As discussed in Sec. \ref{sec4.2}, REFINE reduces GFLOPs by over $3000\times$ compared to rendering-based methods like PUP, yet the wall-clock latency reduction on a high-end GPU (e.g., RTX 3090) is disproportionately smaller ($<20\times$).

This discrepancy stems from differences in hardware utilization. Rendering-based methods saturate GPU capacity through highly parallelized differentiable rasterization, which effectively masks their substantial computational burden. In contrast, REFINE bypasses rendering entirely, relying on lightweight algebraic operations with minimal hardware utilization, which is particularly relevant for deployment on resource-constrained edge devices. \CZADD{LightGaussian also relies on view-wise Gaussian rasterization to accumulate global significance over the training views, although it does not require the backward Fisher computation used by heavier second-order pruning methods.}

\begin{table}[h]
\vspace{-0.5cm}
\centering
\caption{Computation Breakdown for the Entire Scene.}
\setlength{\tabcolsep}{1.5pt} 
\begin{tabular}{@{}lllc@{}} 
\toprule
\textbf{Method} & \textbf{Modules \& Operations} & \textbf{Total FLOPs} & \textbf{GFLOPs} \\ 
\midrule

\multirow{3}{*}{\textbf{REFINE}} 
& $\cdot$ Adaptive Features ($\sim$16) 
& $16 \times N$ 
& \multirow{3}{*}{\textbf{$\sim$1.96}} \\

& $\cdot$ Visibility Subsampling ($\sim$14) 
& $14 \times N \times 64$ 
& \\

& $\cdot$ Intrinsic Coupling ($\sim$23) 
& $23 \times N$ 
& \\

\midrule

\multirow{3}{*}{\CZADD{\textbf{LightGaussian}}}
& \CZADD{$\cdot$ View-wise Rasterization ($\sim$5,000)}
& \CZADD{$5,000 \times N \times |\mathcal{V}|$}
& \multirow{3}{*}{\CZADD{$\sim$3,245}} \\

& \CZADD{$\cdot$ Significance Accum. ($\sim$150)}
& \CZADD{$150 \times N \times |\mathcal{V}|$}
& \\

& \CZADD{$\cdot$ Volume Weighting ($\sim$5)}
& \CZADD{$5 \times N$}
& \\

\midrule

\multirow{3}{*}{\textbf{GHAP}} 
& $\cdot$ Covariance Prep. ($\sim$120) 
& $120 \times N$ 
& \multirow{3}{*}{$\sim$85.0} \\

& $\cdot$ Iterative Clustering ($\sim$40) 
& $40 \times N \times \mathbf{K}$ 
& \\

& $\cdot$ Parameter Decomp. ($\sim$350)
& $350 \times N$ 
& \\

\midrule

\multirow{3}{*}{\textbf{MesonGS}} 
& $\cdot$ Forward Pass ($\sim$5,000) 
& $5,000 \times N \times |\mathcal{V}|$ 
& \multirow{3}{*}{$\sim$9,450} \\

& $\cdot$ Backward Pass ($\sim$10,000) 
& $10,000 \times N \times |\mathcal{V}|$ 
& \\

& $\cdot$ Grad. Accumulation ($\sim$11) 
& $11 \times N$ 
& \\

\midrule

\multirow{3}{*}{\textbf{PUP}} 
& $\cdot$ Diff. Rasterization ($\sim$15,000) 
& $15,000 \times N \times |\mathcal{V}|$ 
& \multirow{3}{*}{$\sim$9,492} \\

& $\cdot$ Fisher Matrix ($\sim$72) 
& $72 \times N \times |\mathcal{V}|$ 
& \\

& $\cdot$ SVD Decomposition ($\sim$500) 
& $500 \times N$ 
& \\

\bottomrule

\multicolumn{4}{@{}l@{}}{
\scriptsize
\textit{*Note: Based on a standard scene with 
$N \approx 2.1 \times 10^6$ primitives and 
$|\mathcal{V}| \approx 300$ views.}
}\\

\multicolumn{4}{@{}l@{}}{
\scriptsize
\textit{$\mathbf{K}$ is GHAP's target block size ($\approx 1000$).}
}

\end{tabular}
\label{tab:Macro_Computation_supp}
\vspace{-0.5cm}
\end{table}

To quantify these efficiency gains, Table \ref{tab:Macro_Computation_supp} provides a detailed complexity breakdown. The total computational cost aggregates primitive-wise operations, scaled by the number of primitives $N$ and training views $|\mathcal{V}|$. Rendering-based methods (PUP and MesonGS) exhibit high complexity that scales linearly with $|\mathcal{V}| \times N$ due to repetitive forward and backward rasterization passes for every view. \CZADD{LightGaussian also scales with $|\mathcal{V}| \times N$, since its global significance score is accumulated through forward rasterization over the training views.} In contrast, REFINE remains remarkably efficient by bypassing rendering entirely and leveraging a geometric formulation independent of view accumulation. The computational costs for the forward and backward passes are estimated as follows:

\noindent \textbf{1) Forward Pass ($\sim$5,000 FLOPs):} The forward pass for a single 3D Gaussian primitive involves four stages: (i) 3D covariance construction from scaling and quaternions ($\sim$100 FLOPs); (ii) 2D covariance projection using the viewing matrix and affine Jacobian ($\sim$150 FLOPs); (iii) view-dependent color evaluation via Degree-3 SH, computing 16 basis polynomials and their dot products with RGB coefficients ($\sim$200 FLOPs); and (iv) rasterization, which evaluates the 2D Gaussian probability density and performs volumetric alpha-blending. Assuming a typical average footprint of $P \approx 150$ active pixels per primitive in standard-resolution scenes, evaluating the quadratic form, exponential, and alpha-blending requires $\sim$30 FLOPs per pixel. This yields an aggregated rasterization cost of $150 \times 30 = 4,500$ FLOPs. Summing these stages, the forward cost is estimated at $\sim$5,000 FLOPs per primitive.

\noindent \textbf{2) Backward Pass ($\sim$10,000 FLOPs):} The backward pass computes loss gradients w.r.t. optimizable parameters (SH, rotation, scaling, opacity, position). Following standard automatic differentiation principles, computing vector-Jacobian products, especially through dense matrix calculus in affine projections and SH polynomials, typically requires $2\times$ to $2.5\times$ the operations of the forward pass. We thus conservatively estimate the backward cost at $\sim$10,000 FLOPs per primitive.

\noindent \textbf{Post-Pruning Fine-Tuning.} Although we evaluated pruning under a strict zero-shot, post-processing setting to directly compare the inherent effectiveness of the importance metrics, REFINE also serves as an exceptionally robust and efficient initialization for subsequent optimization pipelines.

To demonstrate this, we compared REFINE with PUP under a fine-tuning setting at an extreme pruning ratio of 70\%. As shown in Table \ref{tab:fine-tuning_supp}, after removing the redundant primitives, both methods underwent fine-tuning optimization. REFINE achieves consistently higher PSNR across the evaluated scenes, while maintaining highly competitive SSIM. This validates that our rendering-free metric correctly identifies and retains the most structurally significant primitives, providing an optimal starting point for fine-tuning while saving massive computational overhead during the pruning phase.

\begin{table}[h]
\vspace{-0.2cm} 
\centering
\caption{Pruning Comparisons with Fine-tuning (ratio = 70\%)}
  \begin{tabular}{l ccc ccc}
  \toprule
  \multirow{2}{*}{\textbf{Dataset}} & \multicolumn{3}{c}{\textbf{PUP}} & \multicolumn{3}{c}{\textbf{REFINE (Ours)}} \\
  \cmidrule(lr){2-4} \cmidrule(l){5-7}
  & PSNR $\uparrow$ & SSIM $\uparrow$ & LPIPS $\downarrow$ & PSNR $\uparrow$ & SSIM $\uparrow$ & LPIPS $\downarrow$ \\
  \midrule
  bicycle   & 25.01 & 0.745 & 0.244 & \textbf{25.15} & \textbf{0.748} & {0.248} \\
  bonsai & 31.50 & {0.938} & {0.197} & \textbf{31.60} & 0.934 & 0.202 \\
  kitchen  & 29.11 & {0.921} & {0.137} & \textbf{30.55} & 0.913 & 0.152 \\
  \bottomrule
  \end{tabular}
\label{tab:fine-tuning_supp}
\vspace{-0.2cm}
\end{table}
\begin{figure}[h]
  \centering
  \includegraphics[width=0.8\linewidth]{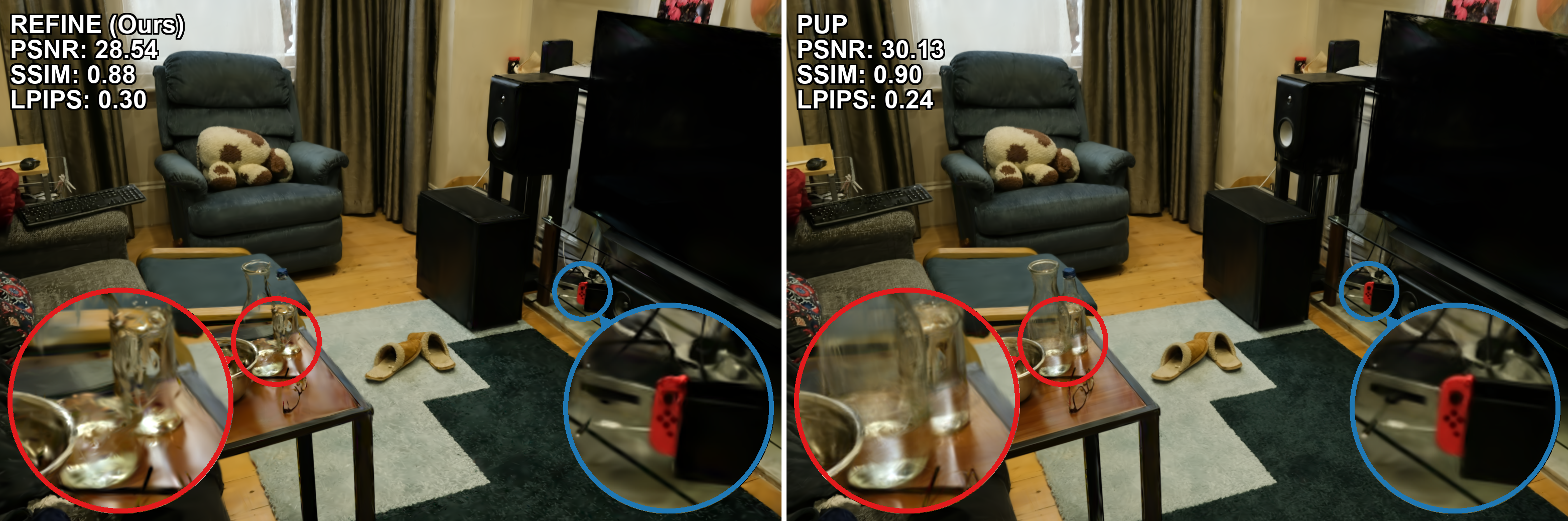}
  \caption{Visual comparisons of a failure case on a very dense scene (Room) with an extreme 70\% pruning ratio. Ignoring cumulative interactions leads to local blurring.}
  \label{fig:failure_cases_supp}
\end{figure}
\noindent\textbf{Analysis of Failure Cases.}
To achieve extreme rendering-free speed, our theoretical formulation introduces two structural approximations: primitive independence and attribute orthogonality. While our empirical results in the main text demonstrate that these assumptions hold remarkably well in general cases, they naturally present limitations under extreme conditions.
Specifically, since our post-processing pruning lacks access to original multi-view images, all current methods rely on non-strict approximations. In highly dense or heavily occluded scenes, and pushed to extremely aggressive pruning ratios (e.g., 70\%), the cumulative cross-primitive interactions, such as alpha blending along a dense ray, become too strong to ignore. As shown in Fig. \ref{fig:failure_cases_supp}, removing primitives based on isolated parameter scores in such densely overlapping areas can amplify the disruption of these interactions, leading to localized blurring and visual artifacts. In these failure cases, our method's PSNR drops slightly below that of fully render-aware methods like PUP. 

Therefore, while REFINE provides a highly effective heuristic metric for ultra-fast pruning, it trades off marginal rendering fidelity in exceptionally dense regions for a $3000\times$ speedup. Future work could explore incorporating lightweight, localized re-rendering passes or low-rank off-diagonal approximations to mitigate this while preserving real-time efficiency.

\end{document}